# Least Entropy-Like Approach for Reconstructing L-Shaped Surfaces Using a Rotating Array of Ultrasonic Sensors

* Nicola Ivan Giannoccaro[1], Giovanni Indiveri[2], Luigi Spedicato[3]

*Department for Innovation Engineering, University of Salento, via per Monteroni, Lecce, Italy*
{[1]ivan.giannoccaro, [2]luigi_spedicato, [3]giovanni.indiveri}@unisalento.it

*Abstract.* **This paper introduces a new algorithm for accurately reconstructing two smooth orthogonal surfaces by processing ultrasonic data. The proposed technique is based on a preliminary analysis of a waveform energy indicator in order to classify the data as belonging to one of the two flat surfaces. The following minimization of a nonlinear cost function, inspired by the mathematical definition of Gibbs entropy, allows to estimate the plane parameters robustly with respect to the presence of outlying data. These outliers are mainly due to the effect of multiple reflections arising in the surfaces intersection region. The scanning system consists of four inexpensive ultrasonic sensors rotated by means of a precision servo digital motor in order to obtain distance measurements for each orientation. Experimental results are presented and compared with the classic Least Squares Method demonstrating the potentiality of the proposed approach in terms of precision and reliability.**

## 1. Introduction

The use of distance sensors is very useful in robotics to support intelligent systems in acquiring awareness of their environment. In particular, ultrasonic sensors have been widely employed due to their robustness and accuracy, especially in those experimental conditions where it is not possible to use other sensors. These sensors make the detection of nearby objects easier although the measured distances may be uncertain especially when specular reflections occur. The possibility of using sonar data for differentiating (i.e. [1]) or reconstructing (i.e. [2],[3],[5],[10]) surface profiles has been discussed in recent contributions.

The authors have lately [7-8] investigated on the ability of the ultrasonic sensors in accurately reconstructing three dimensional (3D) obstacles constituted by two orthogonal planes, where, in the intersection zone, the reflected acoustical waves are subjected to important perturbations as multiple reflections. They used a new type of inexpensive and low frequency sonar sensors able to give out a full waveform signal as well as a Transistor-Transistor Logic (TTL) signal which directly includes information about the Time Of Flight (TOF). They have realized an array composed of four sensors whose beams may be pointed using a controlled digital motor [7]. The sensors are simultaneously fired in order to generate an approximately flat wave front. A new and complex strategy for solving the problem of multiple reflections has been introduced in [7] with a view to reconstructing the obstacle and it has been fully tested and generalized in [8]. This approach firstly considered a parameter derived from the full waveform energy signals; this parameter permits to recognize the angular position corresponding with the intersection of the planes (critical position) [7]. Then, information about the values of the energy-based indicator, the distances and the angular positions were used for plotting a cloud of points [7]. These points were successively passed to the Fuzzy C-Means (FCM) classification function with the aim of partitioning them in three different clusters, one of which contains all the spurious distances. Finally, all the distances relating to each useful cluster were transformed to take into account the direction of reflection and the two sets of polar points (distances and positions) were mapped in a Cartesian reference system. The final points in the Cartesian reference system were then fitted for correctly reconstructing the two planes using the RANdom SAmple Consensus (RANSAC) method.

The RANSAC algorithm is usually employed in computer vision; i.e. in [11] it is used to extract feature points in the three-dimensional space and to calculate the relative pose between the target and the stereo camera. The possibility of using the 3D data points produced by a digital camera is also very interesting in robotics and spacecraft applications [12] for recognizing the interesting areas and the geometry features for the consequence autonomous navigation. Yet camera systems are not able to work in some conditions such as smoky rooms, light-absorbing or shining obstacles. Therefore ultrasonic sensors, also called in-air sonar sensors, are mostly used in these particular scenarios, even if problems like multiple reflections may occur causing an uncorrected reconstruction. For this reason, an important actual research issue in robotics is to test and to introduce automatic and efficient algorithms for using ultrasonic data in feature reconstruction and autonomous navigation.

In this paper a novel and simpler technique is introduced for solving the described reconstruction problem using an ultrasonic scanner; it uses the same preliminary energy-based indicator [8] for partitioning a set of 3D data points into two subsets and, afterwards, a nonlinear function [9] derived from the concept of Gibbs entropy [4] that allows the direct reconstruction of the two planes avoiding the FCM classification, computationally complex, and the following RANSAC algorithm as in [8]. Similar approaches were recently applied [6],[13] on 3D data points acquired by a range camera.

Several experimental tests were carried out for comparing the outcome of the entropy-based method with the direct Least Squares fitting method and with the FCM and RANSAC based approach.

## 2. Least Entropy-Like Estimation: a Brief Overview

For the aim of computing a robust estimate of each scanned plane, the entropy-like estimator is used. The Least Entropy-Like (LEL) method [9] allows to identify the parameters of the planes exhibiting very high robustness to the outlying data erroneously gathered because of multiple reflections. Let $V = \{v_i = (x_i, y_i, z_i)\}_{i=1,...,N}$ be the set of 3D points useful for obtaining the plane defined by (1), where $x, y, z$ are the coordinates of the considered reference system and $\theta = (\theta_1, \theta_2, \theta_3)$ is the *parameter vector*.

$$\theta_1 x + \theta_2 y + \theta_3 z = 1 \tag{1}$$

The error associated to $v_i$ with respect to the model may be expressed as the residual $r_i$ indicated in (2).

$$r_i = v_i \theta^T - 1 = \theta_1 x_i + \theta_2 y_i + \theta_3 z_i - 1 \tag{2}$$

The Least Squares estimation cost $D$ is given by (3) and when it is non-null it is possible to define the *relative squared residuals* $q_i$ as expressed by (4).

$$D = \sum_{i=1}^{N} r_i^2 \tag{3}$$

$$q_i = \frac{r_i^2}{D} \tag{4}$$

The proposed approach is inspired by the definition of Gibbs Entropy. The entropy of a system which admits $N$ discrete states with probabilities $\{p_i\}_{i=1,..,N}$ is computed as follows, where $k$ is a positive constant:

$$\widetilde{H} = -k \cdot \sum_{i=1}^{N} p_i \log p_i. \tag{5}$$

This quantity may be associated to a measure of the distribution of the probabilities such that uniform probability distributions will have maximum entropy while deterministic distributions (namely $p_i=1$ for a fixed $i$, and $p_j=0$ for all $i \neq j$) will have least entropy. The *relative squared residuals* $q_i$ may be considered like the probabilities $p_i$ in the sense that they belong to the interval [0,1] and sum up to 1. Therefore an entropy-like function $H$ may be obtained as in (6):

$$H = \begin{cases} 0 & \text{if } D = 0 \\ -\dfrac{1}{\log N} \sum_{i=1}^{N} q_i \log q_i & \text{if } D \neq 0 \end{cases} \tag{6}$$

that is normalized in the interval [0,1]. There is to notice that the condition $D=0$ is necessary for preventing the singular situation which occurs in case of all null residuals. In fact, when $D=0$ the Least Squares fit is perfect and no further consideration has to be taken. Another aspect of interest is that the entropy like-function also satisfies the properties in (7).

$$\begin{aligned} H &= 0 \quad \text{iff } \exists\,!\ i' : r_{i'} \neq 0 \text{ and } r_i = 0\ \forall i \neq i'; \\ H &= 1 \quad \text{iff } r_i^2 = r_j^2 \neq 0\ \forall i,j \in [1,N] \end{aligned} \tag{7}$$

In physics the configurations of a system with a fraction of highly probable states have lower entropy than those ones where most of states are almost equally probable. This motivation induced to propose a new estimator for the *parameter vector* $\theta$ as in (8). The estimate aims to make the squared residuals as little as possible and the estimator is so based on the minimization of $H$.

$$\hat{\theta}_{LEL} = \arg\min_{\theta} H \tag{8}$$

The LEL method is ideal when the data points include outliers [9] such as ultrasonic data because the devised function only measures the distribution of the relative squared errors without considering the eventually weighted mean square errors. In short, the LEL estimator tends to find a *parameter vector* such that a majority of the data have a small relative squared residual while a small number of them (eventually the outliers) carry a large relative error. For further details refer to [9].

## 3. The Experimental Setup

The purpose of propagating sound waves toward the L-shaped obstacle is achieved by using a mechatronics system consisting of an array of four ultrasonic sensors rotated by means of a digital motor (Fig. 1). The L-shaped structure is built by linking some ply-wooden panels on a frame (Fig. 2).

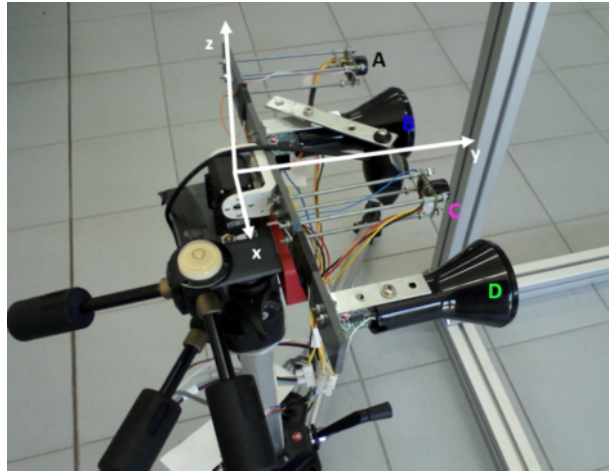

**Figure 1.** The rotating sensors for scanning the surface and the chosen reference system.

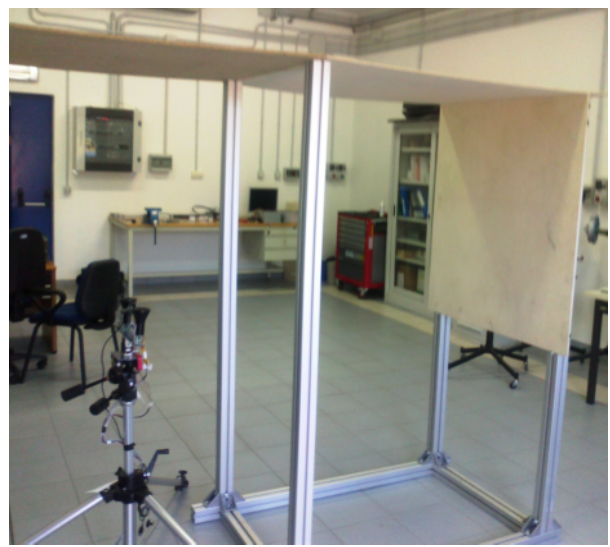

**Figure 2.** The L-shaped obstacle.

The surfaces form a right angle and, when ultrasonic beam are spread toward the cavity due to the intersection, the observed sensor distances are perturbed and associated to outliers. The flat panels are located in such a way that two planes can be fitted: one plane orthogonal to the $y$ axis and the other plane orthogonal to the $z$ axis, where the axes are indicated in Fig. 1. The scanning device is fixed on a professional tripod for leveling and for assuring that the sensors track a portion of a cylinder with axis corresponding to the $x$ axis. The rotating linear array is composed of two types of Hagisonic ultrasonic sensors (AniBat HG-M40DAI and HG-M40DNI) capable to produce and to detect ultrasonic waves at a frequency of 40 kHz. These sensors return an estimation of the TOF. Therefore the distance from the sensor to the obstacle may be evaluated by considering the speed of sound and the characteristics of the media of propagation (the temperature and the humidity of the air). The sensor capsules (indicated with A, B, C, D in Fig. 1) are disposed in an alternate way at a distance of 8 cm one another.

The sensors are rotated by means of a very accurate motor (AX-12 Dynamixel) which is controlled to point the beams at the desired goal position. The sensors are screwed into brackets constrained to the motor shaft. In order to produce a resulting beam characterized by an almost flat wave front, ideal for scanning planes, the sensors were opportunely triggered. Moreover the measurement process is started only after the desired position is reached (motor stopped). Indicating with $\gamma$ the motor angular position around the $x$ axis ($\gamma=0°$ when the capsules lie on the $xy$ plane and the direction of propagation is parallel to the $y$ axis), it is possible to calculate the position of the $i^{th}$ sensor capsules $x^c_{i\gamma}$, $y^c_{i\gamma}$, $z^c_{i\gamma}$ (i=1,..,4) and the sensor distance $d_{i\gamma}$ may be used to determine the data point $\mathbf{v}^r_i = (x_{i\gamma}, y_{i\gamma}, z_{i\gamma})$ by (9).

$$\begin{aligned} x_{i\gamma} &= x^c_{i\gamma} \\ y_{i\gamma} &= d_{i\gamma} \cdot \cos(\gamma) + y^c_{i\gamma} \\ z_{i\gamma} &= d_{i\gamma} \cdot \sin(\gamma) + z^c_{i\gamma} \end{aligned} \qquad (9)$$

The Fig. 3 shows a set of data points resulting from a scanning process. All the calculations and corresponding plots have been obtained by using a technical computer software. It is clear that, in addition to the problems arising at the intersection for multiple reflections, a particular curvature comes to the experimental points in comparison with the real planes. This effect depends on the incorrect hypothesis of direction of reflection corresponding to the pointing direction. Both the effects of multiple reflections and the direction of reflection have been analyzed in [8].

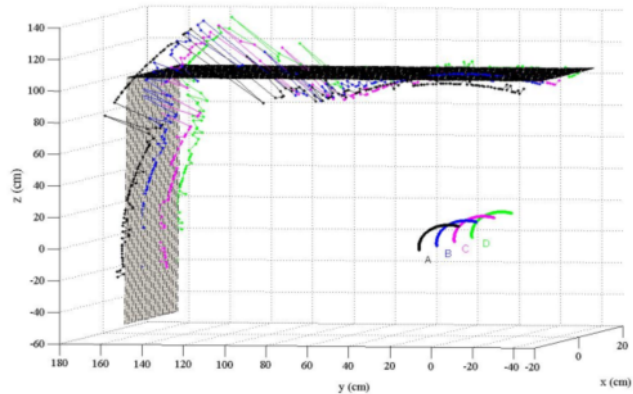

Figure 3. Test 1, experimental data points.

Two experiments will be presented for demonstrating that the final reconstruction is improved by increasing the cardinality of the set of data points. This aim is pursued by programming the motor to make a greater number of rotation steps. The two tests are both referred to the same geometrical setting with the real plane orthogonal to the $y$ axis at a minimum distance of 157 cm and the real plane orthogonal to the $z$ axis at a minimum distance of 106 cm (distances from the origin of the reference system of Fig. 1). The rotation range is [-10° 110°] with 184 steps for the first test (named Test 1) and 220 steps for the second test (named Test 2). For the sake of a correct evaluation, the tripod was equipped by a level gauge and the device position was directly obtained by the smart circuitry. This also allows avoiding eventual problems relating to a low motor torque in some positions. The array rotation was clockwise.

## 4. The Preliminary Analysis

The need of leaving out all the spurious distance values induced the authors to acquire the full waveforms and to value their energy [8]. Let us consider the analogue signal $w_{i\gamma}(t)$ which is returned by the $i^{th}$ sensor in the angular position $\gamma$. The signal energy $E_i^\gamma$ of $w_{i\gamma}(t)$ (in $V^2$s) is defined by (10) where $T$ is the acquisition period ($T=1.8$ s).

$$E_i^\gamma = \int_0^T w_{i\gamma}^2(t)dt \qquad (10)$$

Since the energy trend is similar for each sensor, an indicator of the critical angular position may be derived from the total energy level $E^\gamma$ as in (11).

$$E^\gamma = \sum_{i=1}^{4} E_i^\gamma \qquad (11)$$

The total energy level is plotted in Fig. 4 for Test 1 and it shows that three different peaks may be recognized. Two maximum values are concerned with the direction of direct reflection ($\gamma = 0°$ and $\gamma = 90°$) and another maximum value is relating to the critical position ($\gamma =$ atan(106/157) ≈ 34° in the tests considered).

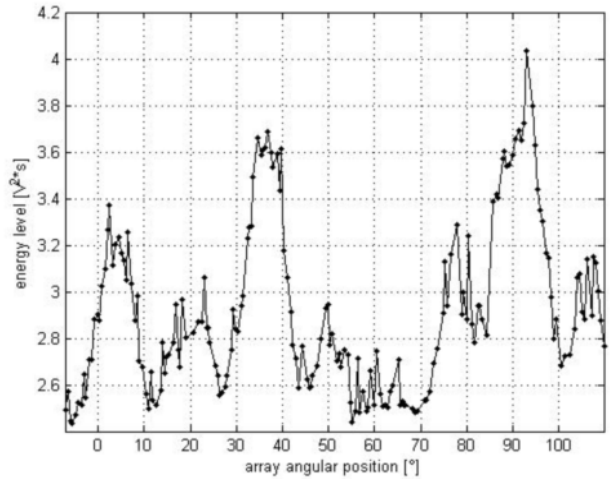

Figure 4. The energy level for Test 1.

The object of automatically identifying this latter peak is attained by means of a theoretical analysis of the expected energy level. The relationship between the total energy value and the distance is achieved by moving the sensor array in front of the L-shaped obstacle when the motor position is $\gamma = 0°$. Considering six different array positions, the points $(d_i, E^{0°}(d_i), i = 1,..,6)$ were used for estimating the model indicated in (12). The distances were evaluated by using a meter [8].

$$E^{0°}(d) = \frac{1}{\varphi_0 d^2 + \varphi_1 d + \varphi_2}, \quad d \in [90,180]\,\text{cm} \qquad (12)$$

The parameters $\varphi_j$ (j=0,1,2) were calculated by the Least Squares Method as in (13). The *Vandermonde matrix* A and the vector B are defined in (14).

$$\begin{pmatrix} \varphi_0 \\ \varphi_1 \\ \varphi_2 \end{pmatrix} = (A^T A)^{-1} A^T B = \begin{pmatrix} -0.00000806 \\ 0.0034138 \\ -0.00193113 \end{pmatrix} \tag{13}$$

$$A = \begin{pmatrix} d_1^2 & d_1 & 1 \\ d_2^2 & d_2 & 1 \\ \vdots & \vdots & \vdots \\ d_6^2 & d_6 & 1 \end{pmatrix}, \quad B = \begin{pmatrix} 1/E^{0°}(d_1) \\ 1/E^{0°}(d_2) \\ \vdots \\ 1/E^{0°}(d_6) \end{pmatrix} \tag{14}$$

The signal energy decreases at increasing the distance for absorption. In fact, the ultrasound energy is converted in other forms which increase the random motion of particles. The maximum sensor to obstacle distance is in critical angular position but the energy gets great around that position, negating (12). As expected, the signal energy is a function of distance and of the incidence angle. This further dependence is due to the fact that the ultrasonic beam is divergent. The vibrating particles cannot transfer all their energy in the pointing direction and this is also the reason that why multiple reflections occur. Let $d_m$ be the median distance of the four distance values gathered in a fixed motor position. The energy level $E^{0°}(d_m)$ may be calculated and all the points $(\gamma, E^{0°}(d_m))$ may be joined to plot the curve of Fig. 5 for Test 1. The ratios R between $E^\gamma$ and $E^{0°}(d_m)$ allow directly determining the critical angular position because the absolute maximum ratio corresponds to it. The ratios are indicated in Fig. 6 for Test 1. They define an optimal indicator of the motor position corresponding to the plane intersection. Such position may be so considered for dividing the 3D data points (shown in Fig. 3 for Test 1) in two subsets.

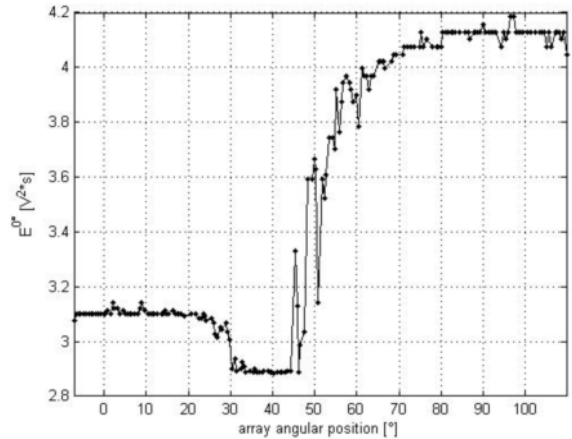

**Figure 5.** The energy level $E^{0°}(d_m)$ for Test 1.

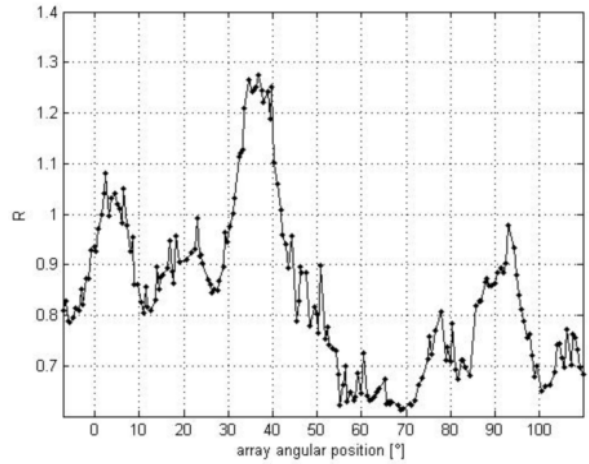

**Figure 6.** The ratios R between $E^{\gamma}$ and $E^{0\circ}(d_m)$ for Test 1.

The four data points relating to the critical position are the unique ones which are excluded; so the two subsets contain all the misrepresented information for multiple reflections. For the sake of avoiding the curvature due to the missing considerations on the direction of reflection, the authors decided to correct the distances before applying (9) [8]. This modification permitted to plot the new data points shown in Fig. 7. The generic point of one of the new subsets may be indicated as $v_i = (x_i, y_i, z_i)$ for estimating each plane.

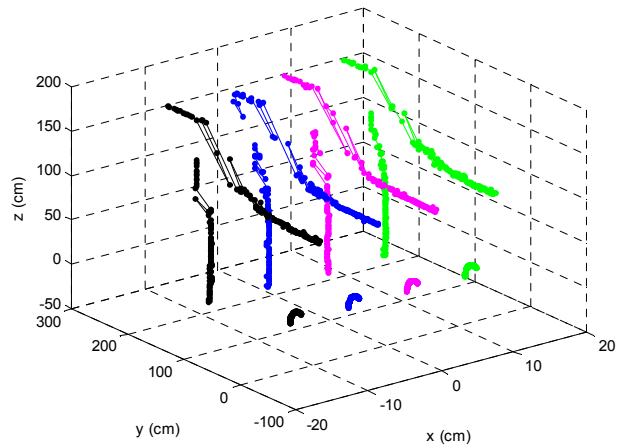

**Figure 7.** Data points for Test 1 after the correction.

## 5. The Experimental Results

### 5.1. Fitting using the Least Squares technique

The points $v_i$ of the two subsets may be named $\{v_i^z = (x_i^z, y_i^z, z_i^z)\}_{i=1,...,M}$ for the plane whose ideal equation is z = 106 (first subset) and $\{v_i^y = (x_i^y, y_i^y, z_i^y)\}_{i=1,...,P}$ for the plane whose ideal equation is y =

157 (second subset). The Least Squares (LS) estimation is obtained by building the matrix *G* having dimensions M×3 for the first subset and P×3 for the second one. Each row of this matrix is given by the coordinates of each data point. The parameter vector may be estimated by (15), where *u* is a column vector of M and P all-ones elements respectively.

$$\hat{\theta}_{LS} = \left(G^T G\right)^{-1} G^T u \qquad (15)$$

The Least Squares fit allows obtaining the equations indicated in (16) for Test 1 and in (17) for Test 2.

$$\begin{aligned} y &= -0.034232x + 0.14493z + 155.4057 \\ z &= 0.014444x + 0.30623y + 104.0506 \end{aligned} \qquad (16)$$

$$\begin{aligned} y &= -0.047382x + 0.11005z + 156.6384 \\ z &= -0.0091748x + 0.28435y + 105.1427 \end{aligned} \qquad (17)$$

An evaluation of the goodness of the reconstruction is given by plotting the planes in the two cases considered (Fig. 8) for Test 1 and Fig. 9 for Test 2).

It is evident from (16), (17) and Figs 8 and 9 that the Least Squares method heavily shows the undesired effects of the multiple reflections and that the reconstructed surfaces are quite different from the real ones. The consequence of an increased number of points for the Test 2 only causes a very small improvement in the results. This aspect is emphasized by comparing the coefficients of *z* in the first equation of (16) and (17) and the coefficients of *y* in the second equation of (16) and (17).

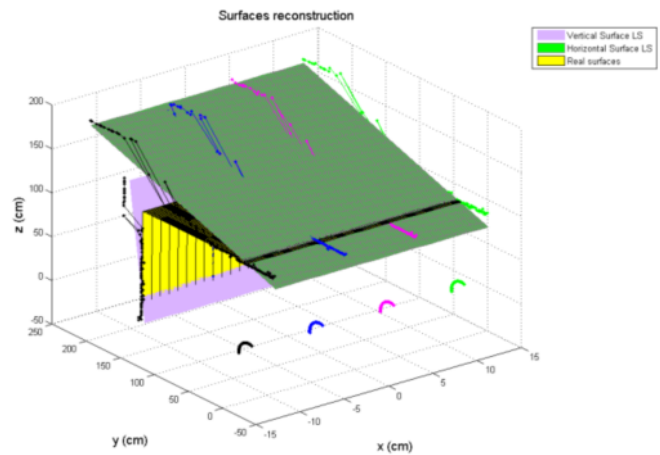

**Figure 8.** Test 1, reconstruction of surfaces and comparison with the real planes applying the LS method.

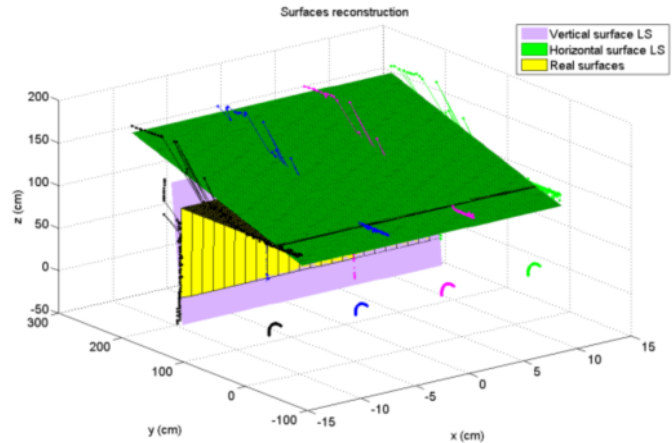

**Figure 9.** Test 2, reconstruction of surfaces and comparison with the real planes applying the LS method.

### 5.2. Fitting using the Least Entropy-Like method

The aim of estimating the parameters of each plane by the LEL method was pursued by calculating the relative squared residuals $q_i$ and defining the entropy-like function $H$ as in (6). According to (8), the function was numerically minimized using a Nelder-Mead algorithm.

As discussed in [9], the LEL estimator is local in nature: hence the numerical minimization of the LEL cost function may be problematic as multiple minima can be present and convergence may thus be affected by the initialization of the minimization routine. Following the analysis and suggestions reported in [9], the numerical minimization of the LEL cost is performed starting from $2 \cdot n$ points in parameter space being n=3 its dimension. The first point is LS solution. Other two points are determined choosing orthogonal vectors to the LS solution with its same norm. The last 3 initialization points are simply opposite to the first 3. Having performed all 6 minimizations, the LEL estimate is chosen to be the one corresponding to the least value of the LEL cost out of the six values.

The LEL fit allows obtaining the equations indicated in (18) for Test 1 and in (19) for Test 2; the relating reconstructed surfaces are plotted in Fig. 10 (Test 1) and 11 (Test 2).

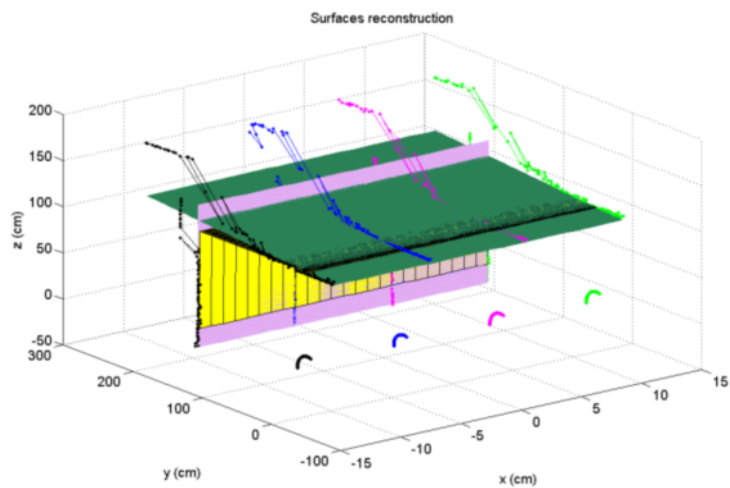

Figure 10. Test 1, reconstruction of surfaces and comparison with the real planes applying the LEL method.

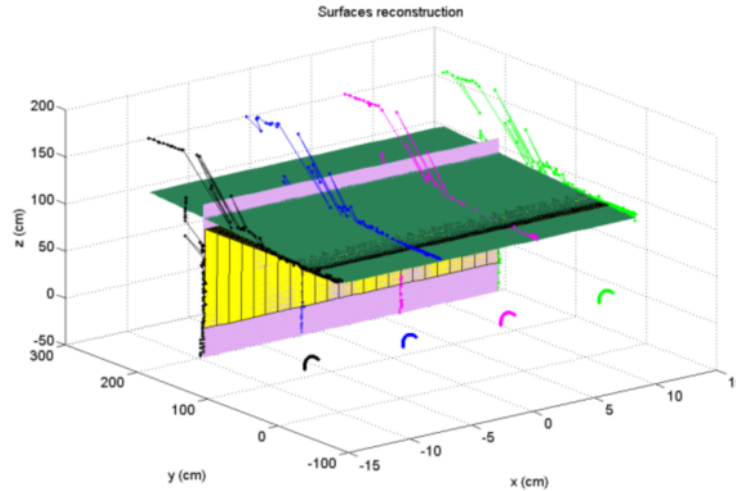

Figure 11. Test 2, reconstruction of surfaces and comparison with the real planes applying the LEL method.

$$y = -0.05493x - 0.002934z + 157.808$$
$$z = 0.017388x + 0.072053y + 105.9031 \quad (18)$$

$$y = -0.02269x - 0.00317z + 157.878$$
$$z = -0.0138x + 0.07046y + 105.8331 \quad (19)$$

It appears clear that the LEL strategy, applied to the subsets identified by the energy indicator, permits to reconstruct the scanned surfaces with a good accuracy; the coefficients of $z$ of the first equation of (18) and (19) decrease of two orders of magnitude with respect to the corresponding ones of (16) and (17); the coefficients of y of the second equation of (18) and (19) decrease of almost one order of magnitude with respect to the corresponding ones of (16) and (17). In spite of the data points affected by the multiple reflections, the Figs 8-11 clearly highlight that the LEL-based reconstruction is much closer to the real configuration than the LS-based fitting.

### 5.3. Further improvements of the LEL method introducing a pre-filtering

In order to test the approximations, the error fitting $e_i$ defined by (20) is calculated for each point $v_i$ of the two subsets.

$$e_i = v_i \cdot \hat{\theta}_{LEL}^T - 1 \quad (20)$$

In Figs 12 and 13 the errors relating to the two subsets are sorted in ascending order. The Median Absolute Deviation (MAD) [9] is also calculated for the four subsets by (21) where $\mu$ is the median error.

$$\text{MAD} = \text{median}(|e_i - \mu|) \qquad (21)$$

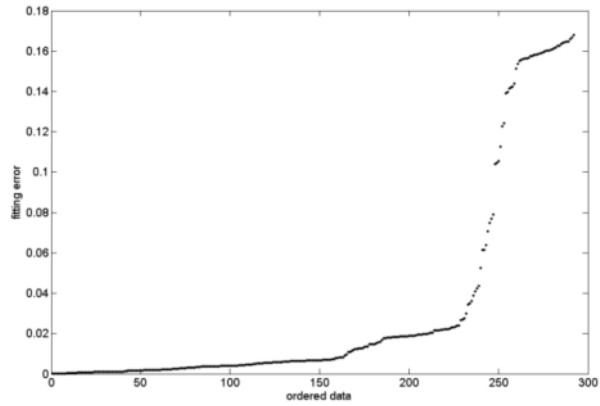

Figure 12. Test 1, fitting error for the estimated horizontal surface.

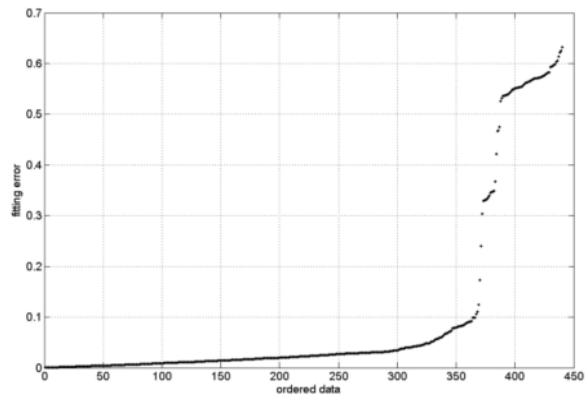

Figure 13. Test 1, fitting error for the estimated vertical surface.

Hypothesizing that the errors are normally distributed, the standard deviation $\sigma$ is calculated as in (22), where $k$ is a constant equal to 1.4826 [9].

$$\sigma = k \cdot \text{MAD} \qquad (22)$$

The standard deviations for Test 1 and Test 2 are indicated in Table 1 ($\sigma^z$ for the horizontal surface and $\sigma^y$ for the vertical one).

Table 1. Standard deviation before filtering.

|  | *Test 1* | *Test 2* |
|---|---|---|
| $\sigma^z$ | 0.0334 | 0.0307 |
| $\sigma^y$ | 0.0098 | 0.0104 |

The fitting error trend suggests removing the data referred to the higher $e_i$ values from the original sets for improving the accuracy of the LEL fit. At this proposal a threshold equal to 10% of the maximum $e_i$ value is imposed for leaving out the data points associated with greater errors.

After filtering, a further application of the LEL strategy allows estimating the surfaces whose equations are expressed by (23) for Test 1 and (24) for Test 2 and plotted in Figs 14 and 15.

$$y = 0.002645x - 0.016042z + 158.2327$$
$$z = 0.062039x + 0.0043015y + 105.4233$$
(23)

$$y = 0.0028592x - 0.014515z + 158.1321$$
$$z = 0.015641x + 0.011899y + 105.5651$$
(24)

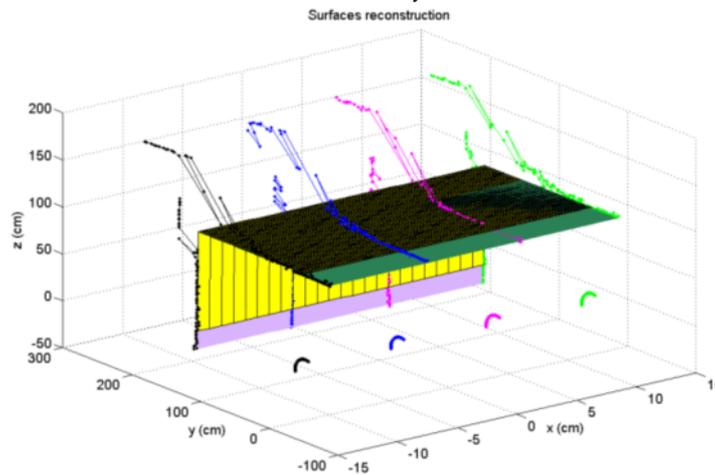

Figure 14. Test 1, reconstruction of surfaces and comparison with the real planes applying the LEL method after pre-filtering.

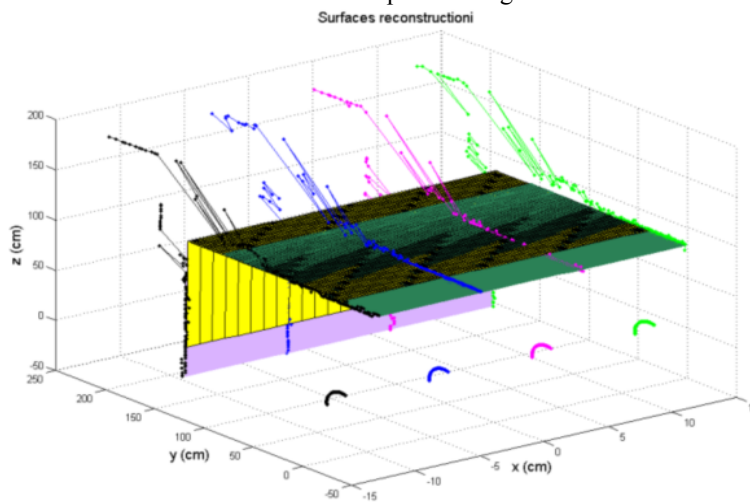

Figure 15. Test 2, reconstruction of surfaces and comparison with the real planes applying the LEL method after pre-filtering.

It is remarkable that pre-filtering improves the final reconstructions. In fact the coefficients of *x* and *z* of the first equations and of *x* and *y* of the second equations decrease and their values come close to zero. The reconstructed surfaces in Figs 14 and 15 are almost overlapped to the real scanned ones.

The new values for the standard deviation are indicated in Table 2. As confirmation of the previous considerations on the utility of the pre-filter, these deviations are lower than the values in Table 1 and they give interesting information on the statistical dispersion.

Table 2. Standard deviation after filtering.

|  | *Test 1* | *Test 2* |
|---|---|---|
| $\sigma^z$ | 0.0264 | 0.0296 |
| $\sigma^y$ | 0.0013 | 0.0018 |

### 5.4. Comparison with the complex strategy based on FCM and RANSAC

A further comparison of the performance of this novel entropy-based strategy is carried out by analyzing the results of the complex approach based on FCM and RANSAC [8] applied to the same experimental data of Test 1 and Test 2. The last approach has been recently introduced for excluding all the misrepresented distances by partitioning the original dataset by means of a soft clustering (FCM). The FCM function permits to obtain three different clusters and the indicator *R* identifies one of them that includes the spurious points. The RANSAC was used for fitting the useful data points concerned with the two remaining clusters.

The outcome of this strategy is expressed in (25) for Test 1 and in (26) for Test 2.

$$\begin{aligned} y &= -0.0167x - 0.0158z + 157.7707 \\ z &= 0.0456x + 0.0341y + 105.7518 \end{aligned} \quad (25)$$

$$\begin{aligned} y &= -0.0418x - 0.0245z + 157.8264 \\ z &= 0.0088x + 0.0246y + 105.5534 \end{aligned} \quad (26)$$

Although the use of these algorithms leads to an accurate reconstruction of the surfaces, there is to note that the entropy method returns good results and it is computationally less complex.

### 6. Conclusions

Many researches are aimed at endowing a robot with sensors in such a way as to let it avoid the obstacles. Lots of them make use of very expensive devices consisting of complex laser systems with a view to reconstructing the surrounding environment. The latter studies allow obtaining very accurate maps and the measurement processes do not need of particular physical considerations. Nevertheless, laser sensors may fail when they operate in some scenarios such as low visibility areas, transparent obstacles or in case of high reflective surfaces. The use of ultrasonic sensors permits to realize a low-cost scanner and some mathematical tools allow easily interpreting and processing the sensory information.

In this paper, the authors propose a new method for processing the ultrasonic data which is based on the minimization of an entropy-like function. The set of data points affected by phenomena of multiple reflections is partitioned by means of an energy indicator. After that two subsets are obtained and transformed for taking into consideration the direction of propagation, the entropy-like estimator is used for appreciating the parameters of the scanned planes. This technique permits to leave out all the outlying data points due to the multiple reflections. For the sake of attributing a confidence to the final reconstructions the equations of the planes for two different tests (characterized by different number of points) are compared with the equations obtained by the Least Squares method and by the strategy based on FCM and RANSAC. All the proposed experiments are referred to a structured scenario, in order to avoid undesired reflections from non-target obstacles. In fact a limit of the present study is concerned with scattering phenomena that may happen in unstructured environments. This aspect will be investigate in future researches. The purpose is to fuse ultrasonic data with other information and to adopt this fresh entropy-based estimator for pattern recognition. We are going to consider the advantage of equipping a rover with the ultrasonic scanner so as to let it perceive the location in its space. The use of ultrasonic proximity sensors will let the robot get out of a maze in case of smoky rooms, light-absorbing or shining obstacles.